\documentclass[cameraready]{Interspeech}

\usepackage{multirow}
\usepackage{float}
\usepackage{booktabs}
\usepackage{comment}

\title{ArFake: A Robust Framework for Multi-Dialect Arabic Speech Spoofing Detection Benchmark}

\author[affiliation={1}]{Mohamed}{Elsetohy}
\author[affiliation={2}]{Alhassan}{Ehab}
\author[affiliation={1}]{Ali}{Mekky}
\author[affiliation={1}]{Besher}{Hassan}
\author[affiliation={3}]{Shady}{Shehata}

\address{
    $^1$ MBZUAI, UAE \\
    $^2$ Queen’s University, Canada \\
    $^3$ University of Waterloo, Canada
}

\email{Mohamed.Elsetohy@mbzuai.ac.ae, shady.shehata@uwaterloo.ca}

\begin{document}
\maketitle

\begin{abstract}

Rapid advances in generative text-to-speech have made it increasingly difficult to distinguish bona fide from synthetic speech, especially for low-resource Arabic dialects, where data scarcity, complex morphology, and non-standard orthography hinder both synthesis and detection. We introduce \textsc{ArFake}, the first end-to-end framework for generating and detecting spoofed Arabic speech across eight dialects. The framework generates speech with four TTS models, then assesses quality and intelligibility using classifier-based measures, downstream ASR performance, and human mean opinion scores. We curate a mixed bona fide/synthetic corpus to train a spoofing detector and evaluate robustness both in-domain and under Leave-One-Generator-Out (LOGO) protocols, achieving 96\% and 97\%. Generalization to unseen dialects is examined with Leave-One-Dialect-Out (LODO) evaluation. \textsc{ArFake} establishes a benchmark and a scalable pathway toward stronger Arabic speech security across generators and dialects.
\end{abstract}

\section{Introduction}
\label{sec:intro}

Audio deepfakes generated by neural TTS and voice-cloning systems can now enable impersonation, voice phishing, and misinformation at scale. While benchmark campaigns such as ASVspoof have driven substantial progress in spoof-speech detection for English, Arabic, specifically its dialect, remains comparatively under-represented, despite its broad dialectal diversity and uneven performance of speech technologies across regions. As generative models continue to advance, the absence of robust Arabic benchmarking resources increases the susceptibility of Arabic-speaking communities to synthetic-speech manipulation.

The central objective of this work is to mitigate a major shortcoming in spoofed-speech detection for low-resource languages, with a particular emphasis on Arabic dialects. Recent progress in generative text-to-speech (TTS) has substantially enhanced speech naturalness, but has simultaneously amplified the threat posed by audio deepfakes and voice spoofing. Despite this growing risk, the anti-spoofing literature remains heavily concentrated on high-resource languages, while Arabic—especially its dialectal varieties—has received comparatively limited attention, even though it is widely spoken and increasingly represented in digital contexts. To address this gap, we propose \textsc{ArFake}, a framework designed for the systematic generation and detection of spoofed Arabic speech.\footnote{\href{https://huggingface.co/datasets/Mohammed01/ArFake}{ArFake Dataset on Hugging Face}}

\noindent\textbf{Contributions.}
\begin{itemize}\setlength\itemsep{2pt}
    \item \textbf{A Comprehensive Arabic Spoofing Pipeline:} We design a five-phase framework covering synthetic speech generation, intelligibility measurement, dataset construction, detector training, and robustness evaluation.
    \item \textbf{Multi-Dialect Arabic Spoofed Dataset:} Using the Casablanca ~\cite{talafha2024casablancadatamodelsmultidialectal} multi-dialectal dataset and four open-source TTS systems, we generate diverse synthetic speech and construct a large-scale mixed bona fide/spoofed dataset designed for Arabic anti-spoofing research.
    \item \textbf{Robust Detectors:} that can distinguish between spoof and bona fide audios.
    \item \textbf{Robustness protocols:} standardized evaluation protocols to support in-domain testing as well as cross-generator (leave- one-generator-out) and cross-dialect (leave-one-dialect-out) robustness.
\end{itemize}

\section{Related Work}

Advances in TTS and voice cloning have dramatically improved synthetic speech realism. Early systems such as WaveNet~\cite{oord2016wavenetgenerativemodelraw} and Tacotron~\cite{wang2017tacotronendtoendspeechsynthesis} required speaker-specific training, while recent zero-shot systems (e.g., VALL-E~\cite{wang2023neuralcodeclanguagemodels}, Voicebox~\cite{le2023voiceboxtextguidedmultilingualuniversal}, XTTS~\cite{casanova2024xttsmassivelymultilingualzeroshot}) can clone voices from seconds of audio. FishSpeech~\cite{liao2024fishspeechleveraginglargelanguage} leverages a large multilingual training scale and has shown strong performance in low-resource settings, including Arabic.

Anti-spoofing benchmarks such as ASVspoof~\cite{wang2024asvspoof5crowdsourcedspeech,yamagishi2021asvspoof2021acceleratingprogress,Nautsch_2021} standardize evaluation (often via EER), but generalization remains challenging, especially on realistic ``in-the-wild'' deepfakes~\cite{müller2024doesaudiodeepfakedetection,Liu_2023}. Partial spoofing presents further difficulties~\cite{Zhang_2023}. Arabic deepfake resources remain limited in scale and dialect coverage~\cite{10154048,10753151}. Recent multi-language datasets such as MLAAD~\cite{müller2025mlaadmultilanguageaudioantispoofing} include Arabic but do not provide broad dialect diversity. Tell me Habibi~\cite{r1} introduces audio-visual deepfake data with Arabic-English code-switching, but targets a different multimodal setting. In parallel, unified detection methods explore improved robustness across diverse attacks~\cite{r2}. Our work targets the gap in \emph{multi-dialect Arabic} anti-spoofing benchmarks by building the first dialectical Arabic framework \textsc{ArFake} and providing benchmark protocols designed to quantify robustness under dialect and generator shifts.

\begin{figure*}[h!]
    \centering
    \includegraphics[width=\textwidth,height=6cm,keepaspectratio]{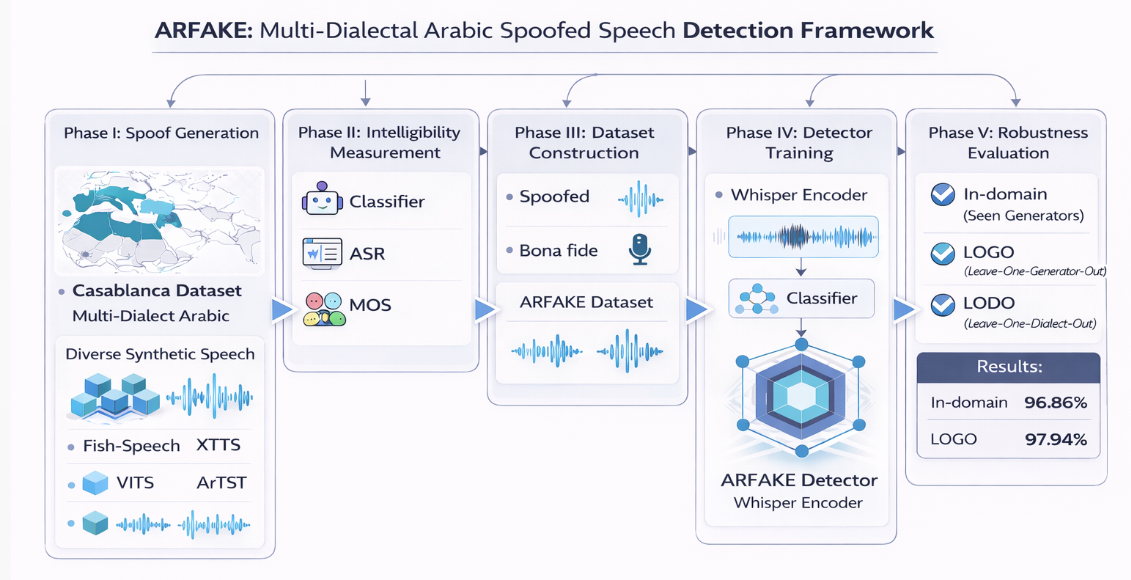}
    \caption{Proposed ARFAKE framework for low-resource Arabic spoofed speech detection, integrating multi-generator synthesis, intelligibility measurement, dataset construction, and cross-generator/dialect robustness evaluation.}
    \label{fig:placeholder}
\end{figure*}

\section{\textsc{ArFake} Framework Methodology}
\label{sec:benchmark}
\label{sec:benchmark}
ARFAKE is an end-to-end framework that systematically addresses Arabic spoofed speech generation and detection across diverse speech generators and dialectal variations.

The ArFake framework methodology is organized into five main stages:

\subsection{Spoof Generation}
The framework uses the Casablanca corpus~\cite{talafha2024casablancadatamodelsmultidialectal}, which contains multi-dialect Arabic speech spanning eight dialect regions (approximately six hours per region). Casablanca provides consistent transcriptions and broad speaker variability, making it suitable for evaluating spoof detection under dialect shift.

For each utterance, we generate spoofed versions using four TTS/voice-cloning systems: XTTS-v2~\cite{casanova2024xttsmassivelymultilingualzeroshot}, FishSpeech~\cite{liao2024fishspeechleveraginglargelanguage}, ArTST~\cite{toyin2023artst}, and VITS~\cite{kim2021conditionalvariationalautoencoderadversarial}. These systems span different training scales and synthesis quality regimes, enabling graded-difficulty evaluation. We treat each generator as a distinct attack family in analysis and robustness testing.


\subsection{Intelligibility Measurement}
To assess the perceptual quality and intelligibility of the spoofed speech generated by the four TTS models, we employed three evaluation methods
\begin{itemize}

\item \textbf{Classifier-based assessment:} was conducted to determine whether the generated utterances could be distinguished from bona fide speech. We used classical classifiers and encoder-based models, providing an initial indication of spoof realism.

\item \textbf{Automatic Speech Recognition (ASR):} was used to evaluate transcription fidelity. We transcribed the spoofed utterances using the Whisper large model~\cite{radford2022robustspeechrecognitionlargescale} and computed Word Error Rate (WER) by comparing the ASR outputs with the original reference transcriptions. It is important to note that the ASR error rate is used as a diagnostic signal and as part of the realism gating process rather than as a standalone measure of perceptual realism.

\item \textbf{Mean Opinion Score (MOS):} We conducted a human perceptual evaluation using the Mean Opinion Score (MOS) to provide a standardized way to quantify subjective human judgments on quality or naturalness. 

\end{itemize}

\subsection{Dataset Construction}
\label{sec:combined}
To construct a robust dataset for spoof detection, we combined bona fide speech with spoofed audio generated by three TTS systems: Fish-Speech, XTTS-v2, and ArTST. Spoofed samples generated by VITS were excluded from the final dataset to be used in our evaluation protocols phase~\ref{sec:rob-pipeline}.
\begin{itemize}

\item \textbf{Training set configuration:} For the training set, 70\% of the bona fide utterances (approximately 13.6k samples) were included. Spoofed samples were incorporated at different proportions per generator: 70\% of Fish-Speech samples, 50\% of XTTS-v2 samples, and 40\% of ArTST samples. This configuration resulted in a training distribution ratio of 30.43\% bona fide and 69.57\% spoofed utterances, ensuring a controlled yet diverse representation of synthetic speech

\item \textbf{Validation split:} From the constructed training data, 10\% was allocated for validation.

\item \textbf{Test set composition:} The test set was formed using the remaining samples, consisting of 30\% of bona fide utterances and 30\%, 50\%, and 60\% of Fish-Speech, XTTS-v2, and ArTST samples, respectively.

\end{itemize}
Overall, the final dataset comprises 54,413 utterances, split into 31,302 samples for training and validation and 23,111 samples for testing.





\subsection{Detectors Training}
After constructing our dataset, we trained the best-performing classifiers identified in Table~\ref{tab:eer_acc_combined}.
\begin{itemize}

\item \textbf{Embedding-based deep learning models:} We used pretrained speech models—HuBERT-base, Whisper-small, Whisper-large, and wav2vec2—to extract high-level speech embeddings capturing semantic and acoustic cues. On top of these embeddings, we added a simple classifier head consisting of two feed-forward layers (each with a fully connected layer, ReLU activation, and dropout) followed by a final classification layer. The models were trained with the Adam optimizer, a learning rate of  $1 \times 10^{-3}$, and a dropout rate of 0.5.


\item \textbf{Traditional machine learning:} A traditional machine learning baseline based on MFCC acoustic features with a Support Vector Machine (SVM) classifier was used. The MFCC-SVM model serves as a conventional baseline for.

\end{itemize}
All models were trained using the constructed dataset~\ref{sec:combined}.

\begin{table*}[t]
\centering
\scriptsize
\renewcommand{\arraystretch}{0.8}
\resizebox{0.8\linewidth}{!}{
\begin{tabular}{@{}lcc|cc|cc|cc@{}}
\toprule
\multirow{2}{*}{\textbf{Model}} & \multicolumn{2}{c|}{\textbf{FishSpeech}} & \multicolumn{2}{c}{\textbf{XTTS-v2}} & \multicolumn{2}{c|}{\textbf{ArTST}} & \multicolumn{2}{c}{\textbf{VITS}} \\
                                & \textbf{EER (\%)$\downarrow$} & \textbf{ACC (\%)$\uparrow$} & \textbf{EER (\%)$\downarrow$} & \textbf{ACC (\%)$\uparrow$} & \textbf{EER (\%)$\downarrow$} & \textbf{ACC (\%)$\uparrow$} & \textbf{EER (\%)$\downarrow$} & \textbf{ACC (\%)$\uparrow$}\\
\midrule
\multicolumn{9}{c}{\textit{Baseline (ASVspoof reference)}} \\
\midrule
RawNet2             & 14.92 & 87.84 & 7.18 & 93.48 & 0.70 & 99.52 & 2.80 & 98.50\\ 
\midrule
\multicolumn{9}{c}{\textit{Embedding-Based Models}} \\
\midrule
Wav2vec2.0       & 43.51 & 63.35 & 2.44 & 98.20 & 0.85 & 99.56 & 0.77 & 99.27 \\
HuBERT-base        & 13.75 & 86.61 & \textbf{0.07} & \textbf{99.96} & 0.08 & 99.93 &  0.84 & 99.52 \\
Whisper-small      & 10.64 & 93.32 & 2.74 & 97.29 & 0.55 & 99.71 & 0.69 & 99.67  \\
Whisper-large    & \textbf{6.92} & \textbf{94.35} & 1.73 & 98.50 & 0.28 & 99.85 & 1.14 & 99.38  \\
\midrule
\multicolumn{9}{c}{\textit{Traditional ML Models (MFCC-Based)}} \\
\midrule
Logistic Regression & 33.29 & 66.73 & 9.44 & 90.29 & 0.16 & 99.74 & 0.23 & 99.78 \\
SVM                 & 14.58 & 85.55 & 3.82 & 96.19 & \textbf{0.00} & \textbf{100} &  \textbf{0.00} & \textbf{100}\\
KNN                 & 37.86 & 57.21 & 8.17 & 88.06 & 0.28 & 99.45 & 0.63 & 99.12 \\
Decision Tree       & 33.67 & 66.84 & 18.35 & 81.25 & 0.47 & 99.63 & 0.91 & 99.23 \\
Random Forest       & 19.56 & 81.07 & 9.18 & 90.73 & 0.07 & 99.96 & 0.10 & 99.93 \\
Gradient Boosting   & 25.13 & 74.58 & 9.91 & 90.07 & 0.08 & 99.96 & 0.15 & 99.71 \\
AdaBoost            & 33.51 & 66.24 & 12.93 & 86.89 & 0.08 & 99.96 & 0.31 & 99.71 \\
Extra Trees         & 20.20 & 80.50 & 7.90 & 91.72 & \textbf{0.00} & \textbf{100} & 0.08 & 99.96 \\
Naive Bayes        & 37.82 & 60.32 & 19.89 & 79.63 & 0.70 & 99.34 & 1.05 & 98.83 \\
\bottomrule
\end{tabular}
}
\caption{EER and accuracy across FishSpeech, XTTS-v2, ArTST, and VITS. Near-perfect results on some generators suggest low-intelligibility artifacts that can inflate performance.}
\label{tab:eer_acc_combined}
\end{table*}

\subsection{Benchmark Protocols (Robustness)}
\label{sec:rob-pipeline}
To ensure a comprehensive and robust evaluation of the proposed ARFAKE detector, we adopt three evaluation protocols:
\begin{itemize}\setlength\itemsep{2pt}
    \item \textbf{In-Domain (seen generators):} In this setting, we evaluate the models using the test split of the constructed dataset described in the Dataset Construction section~\ref{sec:combined}.
    \item \textbf{Cross-Generator (LOGO):} To assess generalization across unseen spoofing techniques, we adopt a Leave-One-Generator-Out (LOGO) protocol. In this setup, the detector is trained on spoofed samples from N - 1 generators and evaluated on the held-out generator. In our case, the held-out generator is VITS, which was excluded from the dataset construction phase. This process evaluates the model’s ability to generalize to previously unseen synthesis methods, which is critical in real-world deployment scenarios where new TTS systems continuously emerge.
    \item \textbf{Cross-Dialect (LODO):} To examine robustness to linguistic variation, we implement a Leave-One-Dialect-Out (LODO) protocol. The detector is trained on seven Arabic dialects and tested on the excluded dialect. This setting evaluates the model’s ability to generalize across dialectal diversity, which is critical for low-resource and multi-dialectal languages such as Arabic

\end{itemize}
Together, these evaluation protocols provide a rigorous and comprehensive assessment of both generator-level and dialect-level generalization, ensuring that the ARFAKE framework is robust to unseen spoofing methods and dialectal diversity.
\section{Experimental Results and Analysis}
\label{sec:results}
\subsection{Metrics and Human Evaluation}
\label{sec:eval}
We report Equal Error Rate (EER) and accuracy (ACC) as the primary evaluation metrics. To diagnose the intelligibility and consistency of the synthesized speech, we compute ASR word error rates (WER) using Whisper-Large, aligning both bona fide and spoofed audio with the same reference transcript.

In addition, we collect Mean Opinion Scores (MOS) from 12 native Arabic speakers using a 1–5 realism scale. For the evaluation, we sample eight utterances per generator, representing one utterance from each dialect. The MOS was then calculated by averaging the ratings across evaluators and samples.

\subsection{\textsc{ArFake} Intelligibility}
\subsubsection{Detector Performance Across Generators}

\label{sec-classifiers}
To evaluate the intelligibility of the datasets generated by each TTS system, we first trained and tested a detector separately on the dataset produced by each system. We treat detector performance as a proxy for how distinguishable the synthesized speech is from genuine speech, i.e., how much the generated distribution diverges from the real one.

As shown in Table~\ref{tab:eer_acc_combined}, the dataset generated by FishSpeech was the most challenging for the classifiers, followed by XTTS-v2, ArTST, and VITS. Notably, Whisper-large achieved the best overall performance on FishSpeech with an EER of 6.92\%, outperforming the other models and improving on the RawNet2 by nearly 7\% EER. HuBERT-base also showed exceptional performance on XTTS-v2, reaching an EER of just 0.07\%.

For the ArTST and VITS datasets (Table~\ref{tab:eer_acc_combined}), nearly all classifiers were able to separate synthesized from genuine speech with near-perfect performance. In particular, SVM achieved 100\% accuracy, while Extra Trees performed similarly, reaching 99.96\% accuracy on the VITS-generated dataset. These results raise an important question: \textbf{\textit{Do such high detection scores truly reflect the quality of the synthesized audio produced by these TTS systems, or do they primarily indicate the presence of easily exploitable artifacts that make the synthetic data trivially separable from real speech?}
 }
 \begin{table}[h!]
\resizebox{0.48\textwidth}{!}{
\begin{tabular}{l|cccccccc|c}
\toprule
 \textbf{TTS} & \textbf{DZ} & \textbf{EG} & \textbf{JO} & \textbf{MA} & \textbf{MR} & \textbf{PS} & \textbf{AE} & \textbf{YE} & \textbf{AVG}\\
\midrule
FishSpeech & \textbf{3.08} & \textbf{4.00} & \textbf{3.83} & \textbf{3.58} & \textbf{4.33} & \textbf{3.92} & 2.92 & \textbf{4.08} & \textbf{3.72}\\
XTTS-v2    & 3.00 & 2.42 & 3.08 & 2.75 & 2.58 & 3.33 & \textbf{3.00} & 3.75 & 2.99\\
ArTST      & 1.75 & 1.42 & 1.83 & 1.83 & 2.25 & 2.25 & 2.08 & 2.00 & 1.93\\
VITS       & 2.17 & 1.08 & 1.33 & 2.42 & 1.92 & 1.33 & 1.50 & 1.83 & 1.70\\
\bottomrule
\end{tabular}}
\caption{MOS (1--5) across eight dialects (ISO country codes). FishSpeech is perceived as most realistic, aligning with its higher detection difficulty.}
\label{tab:calculating_MOS}
\end{table}
\subsubsection{Human MOS: Perceived Realism Across Dialects}

Motivated by the detector results, where FishSpeech and XTTS-v2 produced datasets that were more challenging to distinguish from genuine speech, we conducted a human Mean Opinion Score (MOS) evaluation. As shown in Table~\ref{tab:calculating_MOS}, listeners rated the generated audio on a 1--5 scale in terms of perceived naturalness. FishSpeech attained the highest mean MOS 3.72, indicating that listeners perceived its outputs as most closely matching the speech they commonly encounter in their local context, followed by XTTS-v2 2.99. In contrast, ArTST and VITS were consistently judged to be less natural and realistic, achieving the lowest MOS among the evaluated systems. These results indicate that some TTS systems produce more realistic audio than others. However, an open question remains: \textbf{\textit{How accurate and reliable are the transcriptions produced from speech generated by these models?}}
\subsubsection{ASR-based Intelligibility/Consistency Diagnostic}
\label{ASR-Eval}
\begin{figure}[h]
    \includegraphics[width=0.4\textwidth]{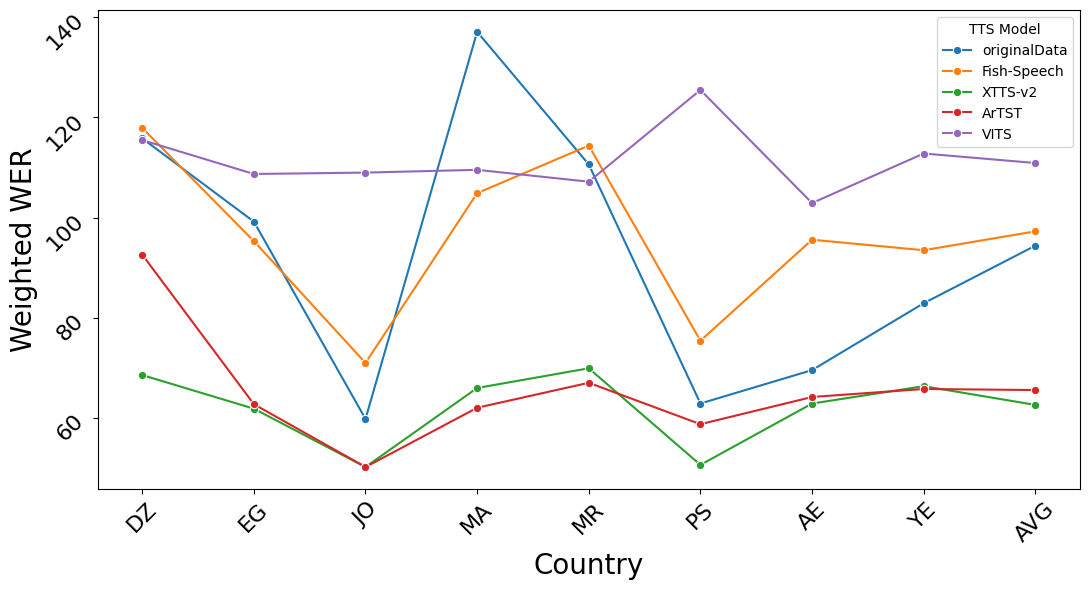}
    \caption{ASR WER across dialects for bona fide and spoofed audio (Whisper-Large). We interpret WER as an intelligibility/consistency diagnostic (and as a component of realism gating), not as a standalone realism measure.}
    \label{fig:WER}
\end{figure}
To further assess the reliability of the transcriptions produced, we evaluated spoofed samples using the Whisper-Large model as described in Metrics Evaluation ~\ref{sec:eval} . For this, the original bona fide data was compared with its ground-truth transcription to establish a baseline weighted word error rate (WER). The spoofed data generated by each TTS model was then aligned against the same original transcription, ensuring a consistent reference for comparison. We report WER, average utterance duration, and perceptual quality. As shown in Figure~\ref{fig:WER}, Fish-Speech achieved an average WER of 97.26\%, which is very close to the original data at 94.40\%, while XTTS-v2 and ArTST reached much lower scores of 62.61\% and 65.58\%, respectively, and VITS obtained 110.90\%, indicating greater ASR difficulty. 

\subsection{\textsc{ArFake} Detectors' Robustness}
Our goal is to build a robust detector that generalizes well across conditions and synthesis methods. However, as TTS systems continue to improve, they will inevitably pose a greater challenge to spoofing detectors. In our experiments, we trained the detector using the training set of the \textsc{ArFake} dataset, as described in Section~\ref{sec:combined}. To evaluate generalization, we assessed the detector's robustness following the protocol outlined in Section~\ref{sec:rob-pipeline}.   
\subsubsection{In-Domain and Out-Of-Domain Generators }
As shown in Table~\ref{tab:eer_acc_combineddataset}, we evaluated our detectors on the \textsc{ArFake} combined test set described in Section~\ref{sec:combined}. Whisper-large achieved the best performance, reaching an EER of 4.88\% on the \textsc{ArFake} test set, followed by Whisper-small with 5.42\%. These results suggest that our detectors are robust to variations in synthesized speech present within the combined evaluation set. However, an important question remains: \textbf{\textit{How do these detectors generalize to synthesis styles that were not observed during training?}}

To examine out-of-domain generalization, we further evaluated the detectors on a fully unseen synthesis style generated by the VITS TTS system, which was excluded from training for this purpose. As the VITS set contains only spoofed samples (i.e., a single class), EER is not defined for this evaluation. Nevertheless, despite not being trained on VITS-generated audio, the detectors still achieved strong performance: Whisper-small obtained 98.30\% accuracy, while Whisper-large followed closely with 97.94\%.

\begin{table}[h]
\scriptsize
\renewcommand{\arraystretch}{0.8}
\resizebox{1\linewidth}{!}{
\begin{tabular}{@{}lcc|cc@{}}
\toprule
\multirow{2}{*}{\textbf{Model}} & \multicolumn{2}{c|}{\textbf{ArFake (combined)}} & \multicolumn{2}{c}{\textbf{VITS (unseen)}}  \\
                                & \textbf{EER (\%)$\downarrow$} & \textbf{ACC (\%)$\uparrow$} & \textbf{EER} & \textbf{ACC (\%)$\uparrow$} \\
\midrule
\multicolumn{5}{c}{\textit{Embedding-Based Models}} \\
\midrule
HuBERT-base        & 7.96 & 96.11 & - & 94.02 \\
Whisper-small      & 5.42 & 96.56 & - & \textbf{98.30} \\
Whisper-large      & \textbf{4.88} & \textbf{96.86} & - & 97.94 \\
\midrule
\multicolumn{5}{c}{\textit{Traditional ML Models (MFCC-Based)}} \\
\midrule
SVM                & 10.76 & 92.08 & - & 73.85 \\
\bottomrule
\end{tabular}}
\caption{EER/ACC on the combined \textsc{ArFake} test set, and accuracy on VITS-generated spoof-only data (unseen during training).}
\label{tab:eer_acc_combineddataset}
\end{table}

\subsubsection{Leave-One-Dialect-Out}
\begin{figure}[h]
  \centering
    \includegraphics[width=0.45\textwidth]{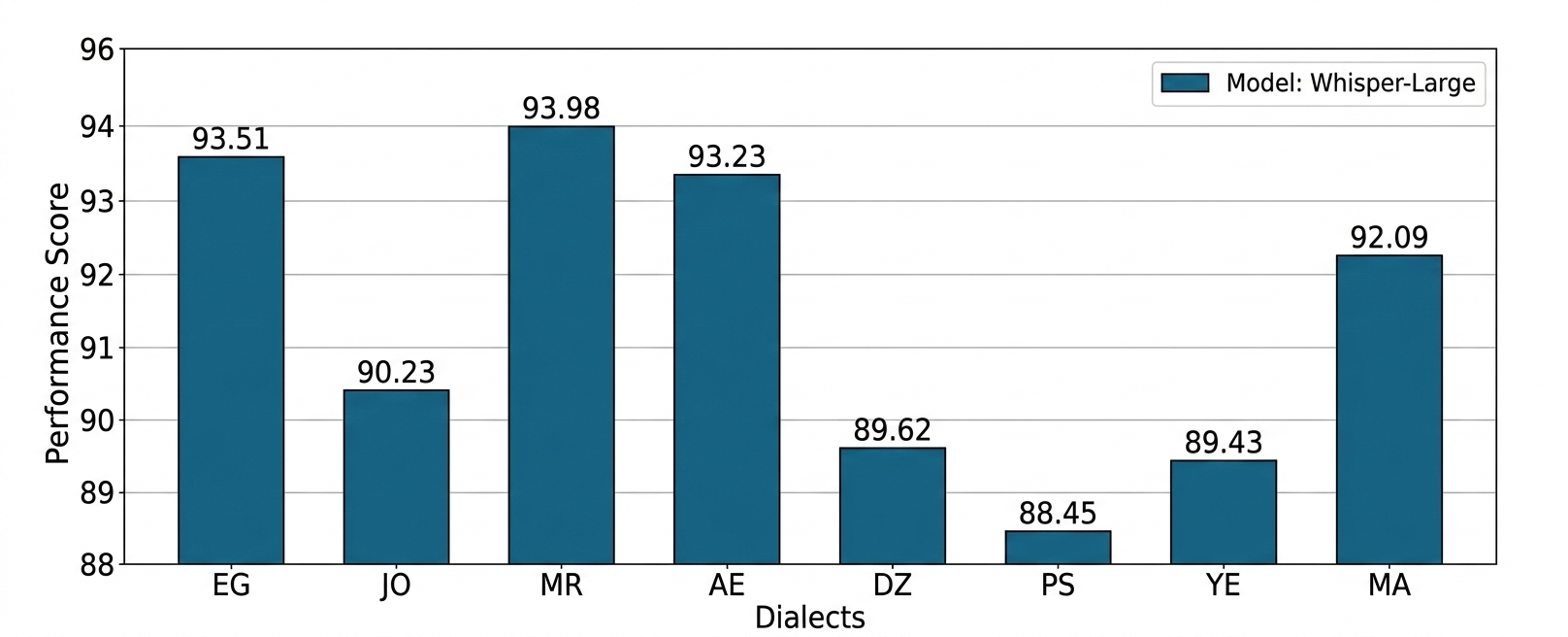}
    \caption{Comparison of Whisper-Large model performance scores across multiple Arabic dialects}
    \label{fig:LODO}
\end{figure}
To assess the detector's performance on unseen Arabic dialects, we trained it following the protocol described in Cross-dialect in Section~\ref{sec:rob-pipeline}. As shown in Figure~\ref{fig:LODO}, the detector generalizes well to held-out dialects, achieving its highest accuracy on the Moroccan dialect (MR) with 93.51\%. The lowest accuracy was observed on the Palestinian dialect (PS), where the detector reached 88.45\%.

\section{Conclusion}
We presented \textsc{ArFake}, the first end-to-end framework for generating and detecting spoofed dialectical Arabic speech. To address the lack of anti-spoofing research for low-resource languages, we designed a five-phase pipeline covering synthetic speech generation, intelligibility measurement, dataset construction, detector training, and robust evaluation. Using multiple TTS systems and the Casablanca multi-dialectal dataset, we built a large-scale Arabic spoofing benchmark and evaluated detection performance under in-domain, (LOGO), and (LODO) protocols. These contributions provide a scalable and reproducible foundation for future research in dialectical Arabic speech security and low-resource language anti-spoofing.

\section{Generative AI Use Disclosure}
Generative AI tools were used exclusively by the authors (a non-native English speakers) to improve language quality, including grammar, style, and clarity. All research contributions—including the study’s conception, literature review, methodology, experiments, data analysis, and interpretation—were developed and executed by the authors, who assume full responsibility for the originality, validity, and integrity of the work.

\bibliographystyle{IEEEtran}
\bibliography{mybib}

\end{document}